\documentclass[preprint,12pt]{elsarticle}

\usepackage{amssymb}
\usepackage{subfigure}
\usepackage{setspace}
\usepackage{geometry} 

\usepackage{multirow}
\usepackage{amsmath}
\usepackage{mathrsfs} 

\usepackage{floatrow}
\newfloatcommand{capbtabbox}{table}[][\FBwidth]
\newcommand{\etal}{\textit{et al.}}
\usepackage{lineno}

\journal{***}

\begin{document}

\begin{frontmatter}



\title{UPDExplainer: an Interpretable Transformer-based Framework for Urban Physical Disorder Detection Using Street View Imagery}

\author[inst1]{Chuanbo Hu}

\affiliation[inst1]{organization={Lane Department of Computer Science and Electrical Engineering, West Virginia University},
            city={Morgantown},
            state={West Virginia},
            country={United States}}

\author[inst2]{Shan Jia\corref{cor1}}
\affiliation[inst2]{organization={University at Buffalo, State University of New York},
            city={Buffalo},
            state={New York},
            country={United States}}

\author[inst3]{Fan Zhang}
\affiliation[inst3]{organization={Department of Civil and Environmental Engineering, The Hong Kong University of Science and Technology},
            city={Hong Kong},
            country={China}}
            
\author[inst4]{Changjiang Xiao}
\affiliation[inst4]{organization={College of Surveying and Geo-Informatics, Tongji University},
            city={Shanghai},
            country={China}}

\author[inst1]{Mindi Ruan}
\author[inst1]{Jacob Thrasher}
\author[inst1]{Xin Li}
\cortext[cor1]{Corresponding author}

\begin{abstract}
Urban Physical Disorder (UPD), such as old or abandoned buildings, broken sidewalks, litter, and graffiti, has a negative impact on residents' quality of life. They can also increase crime rates, cause social disorder, and pose a public health risk. Currently, there is a lack of efficient and reliable methods for detecting and understanding UPD. 
To bridge this gap, we propose \textit{UPDExplainer}, an interpretable transformer-based framework for UPD detection. We first develop a UPD detection model based on the Swin Transformer architecture, which leverages readily accessible street view images to learn discriminative representations. In order to provide clear and comprehensible evidence and analysis, we subsequently introduce a UPD factor identification and ranking module that combines visual explanation maps with semantic segmentation maps. This novel integrated approach enables us to identify the exact objects within street view images that are responsible for physical disorders and gain insights into the underlying causes. Experimental results on the re-annotated Place Pulse 2.0 dataset demonstrate promising detection performance of the proposed method, with an accuracy of 79.9\%. For a comprehensive evaluation of the method's ranking performance, we report the mean Average Precision (mAP), R-Precision (RPrec), and Normalized Discounted Cumulative Gain (NDCG), with success rates of 75.51\%, 80.61\%, and 82.58\%, respectively. We also present a case study of detecting and ranking physical disorders in the southern region of downtown Los Angeles, California, to demonstrate the practicality and effectiveness of our framework.

\end{abstract}






\end{frontmatter}


\section{Introduction}
\label{sec:sample1}

Cities around the world have been grappling with the issue of Urban Physical Disorders (UPDs) for a significant period of time~\cite{sampson1999systematic}. UPD refers to urban landscapes with high levels of decay and deterioration, such as abandoned buildings, litter and trash on streets, broken windows and graffiti, potholes on roads, and overgrown vegetation in public spaces~\cite{seo2018does}. The presence of these environmental disorders not only reduces the quality of life of residents, but also leads to negative outcomes such as increased crime rates, social disorders, and public health problems \cite{ross2001neighborhood, sampson2001disorder, molnar2004unsafe, quinn2016neighborhood, kang2020review}. UPD creates a sense of insecurity, reducing the general livability of urban environments and affecting the well-being of residents \cite{miles2012neighborhood}. For example, a study~\cite{jones2011eyes} reveals that UPD is significantly associated with poorer reading achievement, internalizing behavior problems, and externalizing behavior problems for children. Moreover, UPD is a sign of social and economic decline in urban areas, contributing to a vicious cycle of deinvestment and decay \cite{skogan1992disorder}. Consequently, understanding and addressing physical disorder is essential to create safe, livable, and sustainable urban environments.

Research exploring the impact of UPD on the welfare of individuals and neighborhoods has been of significant interest in the fields of criminology and social epidemiology. Various studies have been conducted to explore and examine the relationships between UPD and specific individual outcomes, such as the behavior of children\cite{jones2011eyes}, emotion and physiological reactivity~\cite{hackman2019neighborhood}, social cohesion and self-rated health\cite{bjornstrom2013social}, suicide mortality~\cite{shen2022exploring}. These studies have emphasized the importance of recognizing and improving UPDs, thus attracting the attention of government bodies to take the necessary action. For example, in light of the potential benefits of greening vacant lots on crime prevention and mental well-being, the US Department of Housing and Urban Development has been actively working to mitigate violent crime through environmental design \footnote{https://www.huduser.gov/portal/pdredge/pdr-edge-frm-asst-sec-082322.html}. This low-cost and scalable solution involves cleaning and greening projects, as well as other interventions such as rehabilitating vacant lots, lands, and deteriorating properties, improving street lighting, and partnering with local law enforcement to assess the built environment and identify vulnerable areas that potentially increase public safety risks. 

In addition to improving UPDs, the identification and comprehension of such disorders are crucial first steps that are in high demand. However, traditional approaches to identifying physical disorder involve conducting physical surveys or relying on citizen complaints~\cite{jones2011eyes,fagan2000street}, which tend to be time-consuming, expensive, and subjective. Furthermore, understanding the reasons and implications of physical disorder requires specific knowledge and expertise, which makes it challenging for policymakers and community leaders to determine practical solutions that address the underlying issues of physical disorder. As a result, there is an urgent need for efficient and reliable approaches that can identify and understand physical disorder in urban environments.

In recent years, street view images have proven to be a reliable tool for observing the microscale built environment, offering a cost-effective alternative to in-person direct observation. 
Researchers have employed street view images to investigate various urban phenomena and applications such as walkability \cite{yin2016measuring}, street greenery quality assessment~\cite{xia2021development}, last-meters wayfinding \cite{hu2023saliency}, 
estimating pedestrian \cite{chen2020estimating}, and built environment \cite{kelly2013using}. Street view images have also provided a valuable and comprehensive resource to detect and analyze physical disorders in urban areas~\cite{nguyen2020using,zhanjun2022multiscale, chen2023measuring}. Meanwhile, rapid advances in machine learning and deep learning techniques have opened up opportunities to extract rich features automatically to analyze street view images. 
 For example, hierarchical deep learning approaches have been developed to detect individual abandoned houses \cite{zou2021detecting}, while deep learning-based methods have been used to identify building facades with graffiti artwork from street view images \cite{novack2020towards}. Recently, Chen \etal \cite{chen2023measuring} designed a deep learning approach to identify various types of physical disorder in street view images and achieved an average accuracy of 72.78\%. 
 
\begin{figure}[t]
 \centering
 \includegraphics[width=1\linewidth]{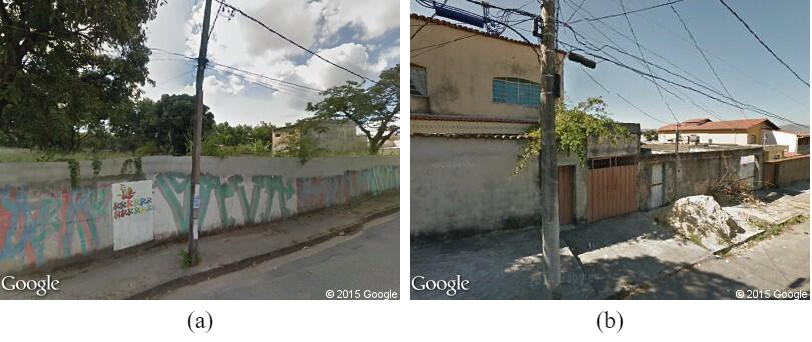}
 \vspace{-0.7cm}
 \caption{Examples of street view images with multiple physical disorder factors, (a) with graffiti wall, messy vegetation, old sidewalk, pole, and old building; (b) with old buildings, dirty walls, pole, and broken sidewalk.}
  \label{fig:0a}
\end{figure}

Existing studies on UPD detection highlight the enormous potential of deep learning-based techniques in measuring and addressing physical disorders in urban environments. However, these methods focused on identifying a particular type of physical disorder factor~\cite{zou2021detecting, novack2020towards, delisle2022deep} or treated a street view image as a scenario of a single-factor disorder~\cite{chen2023measuring}. Taking into account the intricate nature of urban settings, areas characterized by physical disorder often encompass various types of disorder objects (as illustrated in Figure 1). This presents a significant challenge to the precision and effectiveness of current classification methods. Furthermore, while deep learning models have shown promising results in identifying physical disorder factors, the interpretability of these models remains a challenge in the field of computer vision. The ability to explain and interpret the causes and implications of physical disorder in urban environments is also an ongoing research challenge that requires further investigation.

To address the limitations mentioned above in existing studies, this work proposes a new framework called \textit{UPDExplainer} for the detection and explanation of UPDs based on street view imagery and deep learning techniques. Specifically, we first design a Swin Transformer-based~\cite{liu2021swin} model to detect physical disorder in urban environments. Next, we generate score-weighted activation maps from the final convolution layer of the model to provide a visual explanation of the decision-making process. Combining with the semantic segmentation map, the visual explanation map is decomposed to isolate each physical disorder factor and rank the disentangled factors based on their activation map density, namely the contribution to physical disorder detection. This not only offers us a comprehensive understanding of the causes and factors that contribute to UPD, but can also provide valuable guidance to government entities in implementing measures for continuing environmental improvement.

Our work makes the following three contributions to the field of UPD detection.

\begin{itemize}
    \item Detection of generalized UPD. We make the first attempt to employ the cutting-edge Swin Transformer for UPD detection, which aims to learn discriminative and hierarchical visual representations from street view images. Our approach does not exhibit limitations in the type of disorder factor and can be applied to a wide range of physical disorder scenarios. 

    \item Explainable ranking of UPDs. For the first time, we design a UPD identification and ranking module that parses the factors at the semantic level responsible for the physical disorder. This can provide insight into the underlying causes of physical disorders and guide the development of effective interventions. 

    \item Promising performance on real-world data. Experimental results on the public Place Pulse 2.0 \cite{dubey2016deep} dataset with street view images around the world show a detection accuracy of 79.9\% and a top@1 ranking precision of 87.76\%. We also present a case study in the southern region of downtown Los Angeles, California, USA, to demonstrate the effectiveness of our UPDExplainer model. 
\end{itemize}

We hope that the introduction of UPDExplainer, a tool to detect and understand physical disorders in urban settings, will contribute significantly to the advancement of urban disorder analysis. We believe that it will provide reliable and valuable information on its potential applications in urban planning, public policy, and community development.

The remainder of the article is organized as follows. Section 2 offers a brief overview of the relevant literature. In Section 3, we offer a detailed description of the methodology we propose, while Section 4 outlines the experimental design and evaluates the effectiveness of our approach. Section 5 contains an analysis and discussion of our findings, and the paper concludes with a summary in Section 6.

\section{Related Work}\label{sec:2}
In this section, we provide a brief overview of the literature on the detection of physical disorders in urban areas, as well as the public data available to analyze PDU. Following this, we will introduce the relevant research on the interpretability of deep learning models.


\textbf{Detection of urban physical disorder.} 
Existing UPD detection methods can be divided into conventional methods based on human perception and automatic methods using computer vision models. Conventional detection methods typically rely on resident perceptions of their neighborhood's physical and social characteristics~\cite{ross2001neighborhood, mooney2014validity} or use systematic observations by trained researchers~\cite{jones2011eyes, marco2015assessing, franzini2008perceptions}. For example, Jones \etal measured UPD in Los Angeles neighborhoods from structured observations conducted by trained field interviewers. Through an analysis of the connections between disorder and specific individual outcomes, it was discovered that physical disorder is significantly related to neighborhood status and the development of children's reading and behavior. Similarly, Marco \etal~\cite{marco2015assessing} assessed neighborhood disorder based on independent observations by two trained raters who walked 552 census block groups in the city of Valencia (Spain). Several limitations have been noted in these methods, including low detection efficiency based on human resources and perception bias (e.g., resident's assessment can be subjective and sensitive to psychological constructs, stereotypes, and neighborhood prejudices) \cite{marco2015assessing}. 

The second branch of UPD detection methods benefits from the widely used Street View Images (SVI) and rapidly developed deep learning techniques. Specifically, Google street view imagery, as the mainstream open-access data source with global coverage, has served as an aid to provide the physical environment for a variety of urban analyzes~\cite{yin2016measuring,hu2023saliency, chen2020estimating,kelly2013using,hu2020classification}. Rich SVI data, when combined with powerful deep learning in recent years, has
promoted the development of automatic and effective UPD detection. Most studies applied convolutional neural networks to the detection of different UPD factors. For example, Novack \etal \cite{novack2020towards} employed the VGG16 model for the detection of building facades with graffiti. Zhou \etal \cite{zou2021detecting} focused on abandoned house detection and developed a hierarchical deep learning method to leverage both global and local visual features for accurate detection. Similarly, DeLisle \etal \cite{delisle2022deep} used ResNet-50 model to detect abandoned houses. Dubey \etal \cite{dubey2016deep} aimed to quantify urban perception and trained a Siamese-like convolutional neural architecture to learn from joint classification and ranking loss. In \cite{xu2022associations}, Xu \etal ~subjectively and objectively measured street-level perceptions to analyze their relationship with housing prices using computer vision and machine learning techniques. However, these studies concentrated solely on a particular UPD factor \cite{novack2020towards, zou2021detecting, delisle2022deep} or were restricted in their ability to make accurate predictions (e.g., 60\% in \cite{xu2022associations} and 73.5\% in \cite{dubey2016deep}). In contrast, the recent work proposed by Chen \etal \cite{chen2023measuring} made the first attempt to identify various types of physical disorders in street view images and achieved an average accuracy of 72.77\% using the MobileNet V3 model. This study has certain limitations, such as its annotation of the SVI with 15 categories of physical disorders, which may not be sufficient to address the presence of multiple physical disorders in real-world scenarios. Additionally, the classification model lacks interpretability, which hinders a more comprehensive understanding of the causes of UPD.


\textbf{Public data for UPD analysis.}
Well-annotated datasets containing many street view images are essential for UPD detection using deep learning methods. Existing datasets commonly utilize manually labeled data, which is based on how people perceive their environment. For example, Place Pulse~\cite{salesses2013collaborative} is a notable crowd-sourcing effort that aims to map the areas of a city that are perceived as safer, livelier, wealthier, more active, beautiful, and friendly. Users are asked to choose between pairs of images, and as a result, Place Pulse has gathered over 1.5 million reports that evaluate more than 100,000 images from 56 cities. This dataset has been used for early urban perception prediction studies \cite{porzi2015predicting, li2015does}. An updated version called Place Pulse 2.0 \cite{dubey2016deep} is collected based on an online platform from the Massachusetts Institute of Technology (MIT) that enables users to rate the visual appeal of street-level images of different urban neighborhoods. This results in 1.17 million paired comparisons of 110,988 images, originating from 56 cities located in 28 countries spanning 6 continents. A total of 81,630 online volunteers evaluated the dataset based on six perceptual dimensions, including safe, lively, boring, wealthy, depressing, and beautiful. 
Since its launch, Place Pulse 2.0 has been used in various research studies to understand the relationship between the physical characteristics of urban spaces and public perception \cite{zhang2018measuring}. 
Chen \etal \cite{chen2023measuring} constructed a large-scale dataset with 4,876,952 SVIs from 264 Chinese cities for the classification of UPDs. They defined and manually annotated 15 categories of UPD, including abandoned buildings, graffiti, broken roads, messy greening, broken infrastructure, etc.  Since the UPD annotation in this dataset is based on a single factor, it cannot be utilized for UPD detection with complex scenarios and UPD factor ranking analysis. Consequently, we construct our experimental data using the Place Pulse 2.0 dataset in this work.

\textbf{Interpretability of Deep Learning Models.}
Although deep learning techniques have shown significant potential to achieve high accuracy for various complex tasks, it is crucial to understand how the model reaches its conclusions. This has led to a growing interest in developing methods for interpreting deep learning models, with the goal of understanding the reasoning behind their predictions. Such interpretations can help improve the transparency and trustworthiness of the models, as well as identify potential sources of bias or error~\cite{selvaraju2017grad}.

Visual explanations are a class of methods for interpreting deep learning models that use visualizations to highlight the features of the input data that are most important for the model's decision \cite{selvaraju2017grad,wang2020score,ramaswamy2020ablation}. These methods offer a clear understanding of how the model processes the input data and which features are responsible for influencing the output. Visual explanations can also provide guidance to identify instances where the model may be making errors or exhibiting bias. There are various types of visual explanation for deep learning models, such as attention-based maps, activation maps, and occlusion-based methods \cite{selvaraju2017grad, wang2020score}. Attention maps~\cite{zhang2023opti, akhtar2023rethinking} highlight the regions of an image that are most important to the decision of the model by computing the gradients of the output with respect to the input. Activation maps aim to visualize the activations of a model's neurons in response to input, such as Grad-CAM~\cite{selvaraju2017grad}, Score-Cam~\cite{wang2020score}, and Layer-Cam~\cite{jiang2021layercam}, while occlusion-based methods examine the model's response to partially obscured input images~\cite{dong2019explainability}.

Visual explanations have been used in various applications, such as object detection, medical image analysis, autonomous driving, and natural language processing \cite{singh2020explainable, fujiyoshi2019deep, xu2019explainable, ruan2023video}. To our knowledge, there has been no prior research on the interpretation of identifying physical disorder in urban areas. 

\section{Methodology}\label{sec:3}
\begin{figure}[t]
 \centering
 \includegraphics[width=1\linewidth]{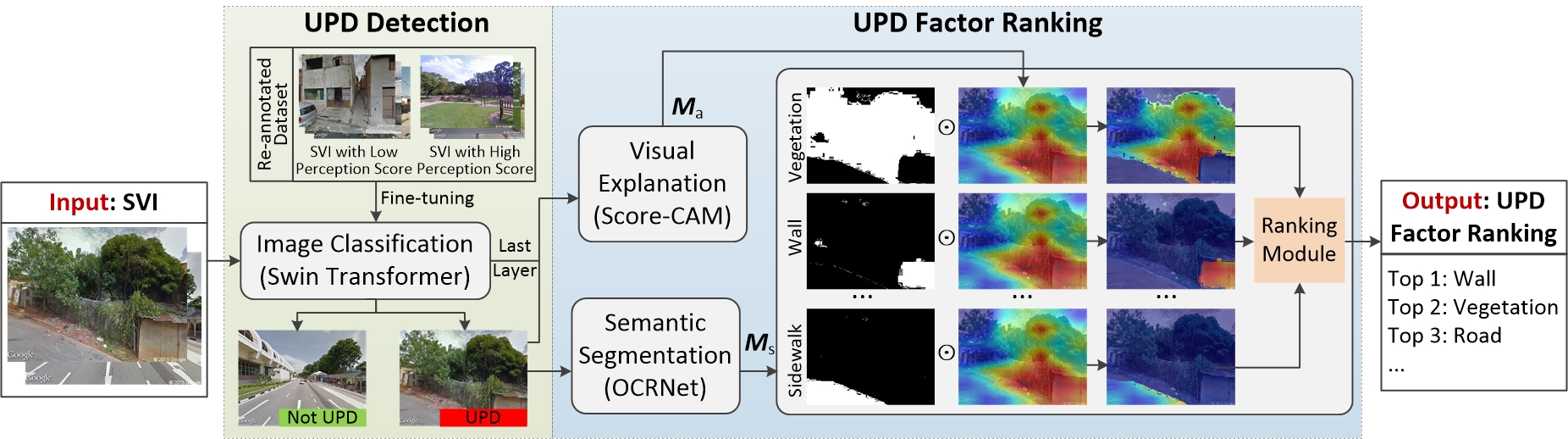}
 \caption{Framework of the proposed UPDExplainer model. }
  \label{fig:1}
\end{figure}

This section provides a detailed description of the proposed UPDExplainer model, which is an interpretable transformer-based framework designed to detect physical disorders in urban environments. We first present the problem formulation to establish a fundamental understanding of our research objective. Next, we introduce the details of the proposed framework, followed by two main components: UPD detection based on the Swin Transformer and semantic object ranking in the UPD scene. The overview of the proposed UPDExplainer framework is illustrated in Figure \ref{fig:1}.

\subsection{Problem Formulation}
The goal of this study is to provide a reliable and efficient approach to detecting and interpreting UPDs. Specifically, we aim to address the following questions:

\begin{itemize}
    \item Can we accurately detect regions of physical disorder in urban landscapes leveraging street view imagery and deep learning techniques?
    \item Would it be possible to identify the particular factors that contribute to the UPD?
    \item Would it be possible to rank the identified factors according to their degree of contribution to the physical disorder?

\end{itemize}

To address these questions, we propose the \textit{UPDExplainer} framework. It first leverages the state-of-the-art Transformer model to acquire comprehensive and distinguishable features for binary UPD detection. In addition, it incorporates a UPD factor ranking module that integrates the semantic segmentation map and the score-weighted activation map, allowing us to identify and rank specific regions and objects strongly associated with physical disorder. 

\subsection{Proposed UPDExplainer Framework}
Given an input SVI image $\boldsymbol{x}$, the UPDExplainer model begins by employing a powerful feature extraction module $f_c(.)$ guided by a binary classification loss as the detection network, which outputs the predicted label $y$, where $y=1$ signifies a UPD image, while $y=1$ indicates a non-UPD image. We then develop the UPD factor ranking module, in which we first extract object regions at the semantic level by applying image segmentation, denoted as $f_s(.)$, to the input image, resulting in the generation of the semantic segmentation map $\boldsymbol{M_s}$. $\boldsymbol{M_s}$ is a binary map where each pixel is assigned a value of either 0 (background) or 1 (foreground/object). Next, we propose to use the visual explanation method $f_v(.)$ in the last layer of the UPD detection module $f_c(.)$, producing a score-weighted activation map $\boldsymbol{M_a}$. This map displays the intensity of contribution from various regions of the image towards UPD detection. The values in the map are normalized to a range between 0 and 1, with 0 being the least significant contribution and 1 being the most significant contribution to the final classification score. Subsequently, we designed a UPD factor ranking function $f_r(.)$ to extract and output the semantic object ranking $r$ based on the merging of the segmentation map and the activation map. Specifically, the whole framework can be presented as
\begin{equation}
    y = f_c(\boldsymbol{x})
\end{equation}
\begin{equation}
    r = f_r(f_s(\boldsymbol{x}), f_v(f_{cla}(\boldsymbol{x}))) = f_r(\boldsymbol{M_s}, \boldsymbol{M_a})
\label{equ:equ2}
\end{equation}
where $f_{cla}$ means the extraction operation of the feature map from the last-layer attention map of the $f_c(.)$ function. 
This framework has several potential applications, including monitoring the disorder neighborhoods of urban areas, identifying areas that require maintenance and renovation, and assessing the effectiveness of urban revitalization efforts. By automating the detection of physical disorders, this framework can help urban planners and policymakers make more informed decisions about the allocation of resources and the design of urban environments.

\subsection{Urban Physical Disorder Detection Based on Swin Transformer}

Detecting UPD with precision using street view images and deep learning models is a topic that has not yet been thoroughly explored. In this paper, we develop a novel approach to the detection of physical disorders in urban settings that utilizes the Swin Transformer \cite{liu2021swin} architecture as $f_c(\boldsymbol{x})$, which is a state-of-the-art computer vision backbone with outstanding performance in a wide range of vision tasks, including image classification and object prediction tasks. It builds upon the popular Transformer architecture originally proposed for natural language processing. The Swin Transformer uses hierarchical feature maps to improve the efficiency of processing large images, and also incorporates a shifted window mechanism that allows for efficient and effective modeling of long-range dependencies in image data. Using the Swin Transformer architecture, our detection model can not only capture distinctive features that can differentiate UPDs from tidy environments, but also exhibit generalized performance in various street view image scenarios.  

\begin{figure}[t]
 \centering
 \includegraphics[width=1\linewidth]{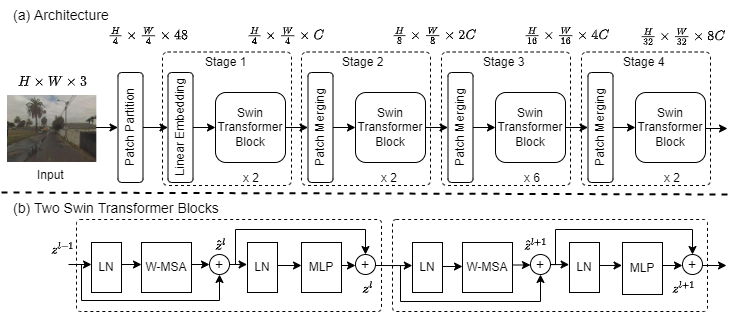}
 \caption{Model structure of Swin Transformer.}
  \label{fig:3}
\end{figure}

The Swin Transformer architecture is outlined in Figure \ref{fig:3}. The SVI input image $\boldsymbol{x}\in \mathbb{R}^{H \times W \times 3}$ (where $H$ and $W$ are the height and width of the image, respectively) is initially segmented into non-overlapping patches using a patch splitting module. Each patch is then considered a ``token", and its feature is established as the raw RGB values of the pixels concatenated together. For our implementation, a patch size of $4 \times 4$ is utilized, resulting in a feature dimension of $4 \times 4 \times 3 = 48$ for each patch. Subsequently, a linear embedding layer is used to project this raw value feature onto an arbitrary dimension (denoted by $C$). Using modified self-attention computations, a number of Swin Transformer blocks are employed to process the patch tokens. The transformer blocks preserve the initial number of tokens, which is calculated as $\frac{H}{4} \times \frac{W}{4}$, and when combined with the linear embedding, they constitute stage 1. To generate a hierarchical representation, the network employs patch merging layers that reduce the number of tokens as the network becomes deeper. The first patch merging layer concatenates the features of $2 × 2$ neighboring patch groups and applies a linear layer to the resulting 4C-dimensional concatenated features. This results in a reduction of tokens by a multiple of $2 \times 2 = 4$ (equivalent to a 2× downsampling of resolution), and the output dimension is set to $2C$. Following this, the Swin Transformer blocks are used for feature transformation, with the resolution maintained at $\frac{H}{8} \times \frac{W}{8}$. The patch merging and feature transformation block that occurs initially is denoted as Stage 2. This step is then repeated twice, leading to Stage 3 and Stage 4, with output resolutions of $\frac{H}{16} \times \frac{W}{16}$ and $\frac{H}{32} \times \frac{W}{32}$, respectively. 

The Swin Transformer architecture replaces the standard multihead self-attention (MSA) module in a transformer block with a module based on shifted windows, while keeping the other layers unchanged. As depicted in Figure \ref{fig:3}(b), a Swin Transformer block comprises a shifted window-based MSA module (W-MSA), followed by a 2-layer Multilayer Perceptron (MLP) with Gaussian Error Linear Unit (GELU) \cite{hendrycks2016gaussian} nonlinearity in between. LayerNorm (LN) layers are applied before each MSA module and each MLP, and a residual connection is applied after each module. The self-attention module in the Swin Transformer operates within local windows that are evenly partitioned across the image in a nonoverlapping manner. If each window contains $M \times M$ patches, the computational complexity of a global MSA module and a window-based module on an image of $h \times w$ patches can be determined following the definitions in~\cite{liu2021swin}:

\begin{equation}
  \Omega \left ( MSA \right ) = 4hwC^{2} +  2(hw)^{2}C
\end{equation}
\begin{equation}
  \Omega \left ( W-MSA \right ) = 4hwC^{2} +  2M^{2}hwC
\end{equation}

\noindent where the computational complexity of the global $MSA$ module is quadratic to the patch number $hw$, whereas that of the window-based module is linear when the window size $M$ is fixed (defaulted to 7). With a large $hw$, the use of global self-attention computation is generally unfeasible, while window-based self-attention remains scalable. 

The approach to partitioning windows in Swin Transformer blocks involves alternating between two configurations. In the first module, a regular window partitioning strategy is used, where the $8 \times 8$ feature map is evenly partitioned into $2 \times 2$ windows of size $4 \times 4$ ($M=4$) starting from the top-left pixel. The next module utilizes a windowing configuration that differs from the previous layer by shifting the window positions by (2, 2) pixels from the regularly partitioned windows. 

Using the above Swin Transformer as a backbone, our detection model is trained on a large dataset of street view images, annotated with labels indicating the presence or absence of physical disorder. The trained model can then be used to automatically detect physical disorders in diverse urban images, enabling faster and more efficient detection of UPDs.

\subsection{Semantic UPD Factor Ranking}

Although detection of PD plays a critical role in uncovering physical disorders in urban settings, existing studies often overlook the importance of providing guidance for subsequent improvement in PD. As different factors can contribute to this disorder, the associated costs of addressing them vary.
For example, a trash pile on the sidewalk and an abandoned building can contribute to physical disorder in a particular area. The cleaning of the sidewalk is often less expensive than the demolition of abandoned buildings. Therefore, it is of utmost importance to distinguish between different semantic objects on the street (e.g., sidewalk, building, wall, and vegetation) related to UPD considering its significant role in the future reconstruction of the urban street environment and UPD governance.

To fill this gap, our UPDExplainer proposes to take the Swin Transformer-based UPD detection as the first step, which will also provide guidance for the next UPD identification and ranking module. Specifically, we have designed the UPD factor ranking $f_r(x)$ to provide fine-grained identification and the ranking of semantic objects based on their contribution to UPD detection. This module involves the following three steps. 
\begin{itemize}
    \item [1)] \textbf{Semantic Segmentation Map Extraction.} The semantic segmentation map $\boldsymbol{M_s}$ is extracted from the UPD street view image using the image segmentation module $f_s(\boldsymbol{x})$ based on OCRNet~\cite{YuanCW20}, which is pre-trained on the Cityscape dataset~\cite{cordts2016cityscapes} containing complex street view images. This module allows us to identify all specific objects included in the UPD street view image. We consider 12 categories of semantic objects covered in the Cityscape dataset~\cite{cordts2016cityscapes}, including sidewalk, building, vehicle, fence, motorcycle, person, pole, road, sky, traffic sign, vegetation, and wall. 
    \item [2)] \textbf{Visual Explanation Map Extraction.} We introduce a visual explanation module $f_v(x)$ to provide evidence of our UPD detection process. It is based on the Score-CAM~\cite{wang2020score} model, which is a state-of-the-art score-weighted visualization method that highlights the regions of the image most strongly associated with UPDs. We apply Score-CAM to the last layer of the Swin Transformer model for the generation of $q$ visual attention maps $\boldsymbol{M_a}$. On this map, warmer colors such as red, orange, and yellow indicate higher activation levels, that is, a larger contribution to the detection of UPD, whereas cooler colors such as blue and green denote lower activation levels. Using this approach, we can gain valuable insight into the factors that contribute to disorder and prioritize interventions that can have the most significant impact on improving quality of life in urban environments. 
    \item [3)] \textbf{UPD Factor Ranking.} A UPD factor identification and ranking module $f_r(.)$ is designed to obtain the semantic object ranking $r$ following Equation (\ref{equ:equ2}). By analyzing the density of each semantic object within the visual attention map, we can determine the top objects that are most closely associated with physical disorder. This process is defined as 

\begin{equation}
    r = f_r(\boldsymbol{M_s}, \boldsymbol{M_a}) = sort([i, \frac{\sum_{j=1}^{N_i}{M_{s_{ij}}M_{a_{ij}}}}{N_i}])
\end{equation}
\end{itemize}

\noindent where $sort(.)$ represents sorting each semantic object $i$ based on the contribution values to UPD in descending order, 
while $N_i$ represents the total number of pixels in the semantic object $i$. $M_{s_{ij}}$ is the semantic area associated with that object $i$, while $M_{a_{ij}}$ represents the value of the visual attention map for that object $i$ at a particular pixel location $j$.

In general, our approach represents a significant step forward in the development of computer vision tools to detect and understand physical disorders of the urban environment. It has the potential to inform and guide targeted interventions aimed at improving the quality of life in urban environments.


\section{Experiments}\label{sec:4}

This section presents extensive experiments that assess the effectiveness of the UPDExplainer model. We will begin by introducing the experimental settings used to implement the proposed framework. Following this, we perform comparative experiments to demonstrate the UPD detection performance of our UPDExplainer, and then evaluate the performance of the UPD ranking module. Subsequently, we evaluate the sensitivity of the proposed method to different semantic object segmentation methods, visual explanation methods, and street view scenarios. To demonstrate the practicability of the model, we conclude by presenting a case study that applies the UPDExplainer to the southern region of downtown Los Angeles, California, USA.

\subsection{Experimental Settings}\label{sec:4.1}

\begin{itemize}
\item \textbf{Dataset.} Our dataset comes from Place Pulse 2.0 \cite{dubey2016deep}, which includes 110,988 street view images captured between 2007 and 2012, covering 56 cities in 28 countries on six continents. Human perceptual ratings were obtained using an online platform where participants were asked to rate pairs of images based on different attributes, including ``safe", ``beautiful", ``lively", ``wealthy",  ``depressing" and ``boring". We re-annotated this dataset based on the attribute rating for our UPD detection tasks. First, we extracted all images that have been rated by humans more than 100 times. Second, we calculate the perception score of each attribute by calculating the Q score \cite{zhang2018measuring}. Third, we extracted approximately 5,000 street view images with a low perception score considering the above six attributes as UPD images. These images were selected from the bottom 5\% of the chosen SVI in the Place Pulse 2.0 dataset.
Furthermore, we randomly selected 5,000 street view images with a high human perception score at the top 95\% as non-UPD images. To evaluate the performance of the proposed method in ranking UPD semantic objects, we manually labeled approximately 500 street view images with UPD by providing the ranking of semantic objects that contribute to UPD.


\item \textbf{Model Training.} The proposed UPDExplainer is trained with the cross-entropy loss for the UPD detection module. Specifically, we fine-tuned the Swin Transformer architecture using our self-annotated street view imagery dataset. The training-validation split was set at 7:3, and we trained the model for 200 epochs, employing the Adam optimizer with a learning rate of 1e-4.

\item \textbf{Evaluation Metrics.} We first use several evaluation metrics to report the UPD detection performance of our approach, including Accuracy, Precision, Recall, and F1-score. Next, we report three metrics to show the UPD ranking performance following previous ranking-based studies ~\cite{taylor2008softrank,efron2011estimation}. They are 1) mean Average Precision (mAP) which measures the average precision of the top-$k$ ranked semantic objects for different values of $k$; 2) R-Precision (RPrec) which evaluates the effectiveness of semantic object ranking by measuring the precision of the top-R ranked objects; and 3) Normalized Discounted Cumulative Gain (NDCG) which measures the quality of the ranking by considering the relevance of each semantic object.

\item \textbf{Hardware and Software Configuration.} All experiments were carried out on a workstation with an Intel Core i7 processor and three NVIDIA GeForce RTX 2080 Ti GPUs. We used Python 3.8 and the PyTorch deep learning framework for model training and evaluation.

\end{itemize}


\subsection{Comparison of UPD Detection}\label{sec:4.2}

\begin{table}[t]
\begin{tabular}{l|c|c|c|c}
\hline
Methods & Accuracy & Recall & Precision & F1\\
\hline
Novack \etal \cite{novack2020towards} & 72.53\% & 71.06\% & 73.15\% & 72.26\%       \\
\hline
DeLisle \etal \cite{delisle2022deep}& 71.93\% &  72.26\% &    71.73\% &   71.99\%  \\
\hline
Chen \etal \cite{chen2023measuring} & 73.35\%   &  68.57\% & 75.76\% & 71.99\%      \\
\hline
VIT \cite{dosovitskiy2020image}& 76.02\% & 73.48\% &  77.40\%   &   75.39\%   \\
\hline
UPDExplainer (ours) & \textbf{79.89\%} &  \textbf{75.87\%} & \textbf{82.48\%}  &      \textbf{79.04\%}  \\
\hline
\end{tabular}
\caption{Comparison of model performance in detecting UPD with the baseline models.}
\label{tab:1}
\end{table}

In this section, we present the experimental results of our proposed approach in UPD detection, and compare it with several baseline detectors.

We consider four baseline models for UPD detection. Three of them, namely the Novack \etal \cite{novack2020towards} method using the VGG16 model \cite{simonyan2014very}, DeLisle \etal \cite{delisle2022deep} method with the ResNet50 model \cite{he2016deep}, and Chen \etal \cite{chen2023measuring} using MobileNet-based model \cite{howard2017mobilenets}, have been used before in this context. The fourth model is another transformer-based approach known as VIT \cite{dosovitskiy2020image}.
All methods were pre-trained on ImageNet \cite{deng2009imagenet} and fine-tuned on our street view imagery dataset as a binary classification task to predict the presence or absence of UPDs.

Table \ref{tab:1} shows that our proposed approach based on the Swin Transformer outperforms all baseline models in terms of accuracy, precision, recall, and F1 score. Specifically, our approach achieves an accuracy of 79.89\%, which is 3.87\% higher than the best-performing baseline VIT model. 

\begin{figure}[t]
 \centering
 \includegraphics[width=1\linewidth]{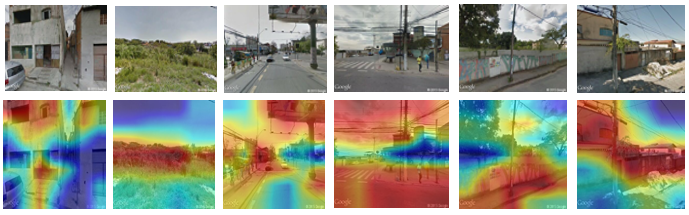}
 \caption{Visualization of UPD detection of the proposed UPDExplainer model. Note that warmer colors (e.g., red, orange, and yellow) indicate higher activation levels, namely a larger contribution to the UPD detection, whereas cooler colors (e.g., blue and green) denote lower activation levels.}
  \label{fig:4}
\end{figure}

Furthermore, we present the visual explanation maps of Score-CAM from our detection model in Figure \ref{fig:4}. The results demonstrate the effectiveness of the Swin Transformer for the task of detecting UPD based on street view imagery. Regions with objects of physical disorder, such as broken houses, messy vegetation, old roads, and poles, can be accurately highlighted by warmer colors, which means a high-intensity activation level for the UPD decision.

\subsection{Evaluation of UPD Factor Ranking}
\label{sec:4.3}

This section presents an experimental evaluation of the proposed UPDExplainer in UPD factor ranking. The ranking metrics are reported for Top@1, Top@2, Top@3, and Top@4, indicating the model's ability to correctly identify one, two, three, and four factors, respectively. Table \ref{tab:2} shows that the proposed framework achieves a mAP, RPrec, and NDCG of 87.76\% for the Top@1 ranking, indicating that the pipeline can accurately rank semantic objects according to their contribution to the prediction of physical disorder. It is not surprising that performance decreases as the number of ranked factors increases, as observed when comparing different rows in Table \ref{tab:2}.
\begin{table}[h]
\begin{tabular}{l|c|c|c|c}
\hline
Metrics  & Top@1 & Top@2 & Top@3 & Top@4\\
\hline
mAP & 87.76\%  & 82.65\% &  78.23\% &  75.51\%   \\
\hline
RPrec & 87.76\% & 84.69\% &  82.31\% &  80.61\%  \\
\hline
NDCG & 87.76\% & 85.38\% & 83.55\% & 82.58\%      \\
\hline
\end{tabular}
\caption{Performance of UPD factor ranking of the proposed UPDExplainer model. Note that the ranking list in the final column corresponds exactly to the ground-truth ranking.}
\label{tab:2}
\end{table}

\begin{figure}[t]
 \centering
 \includegraphics[width=1\linewidth]{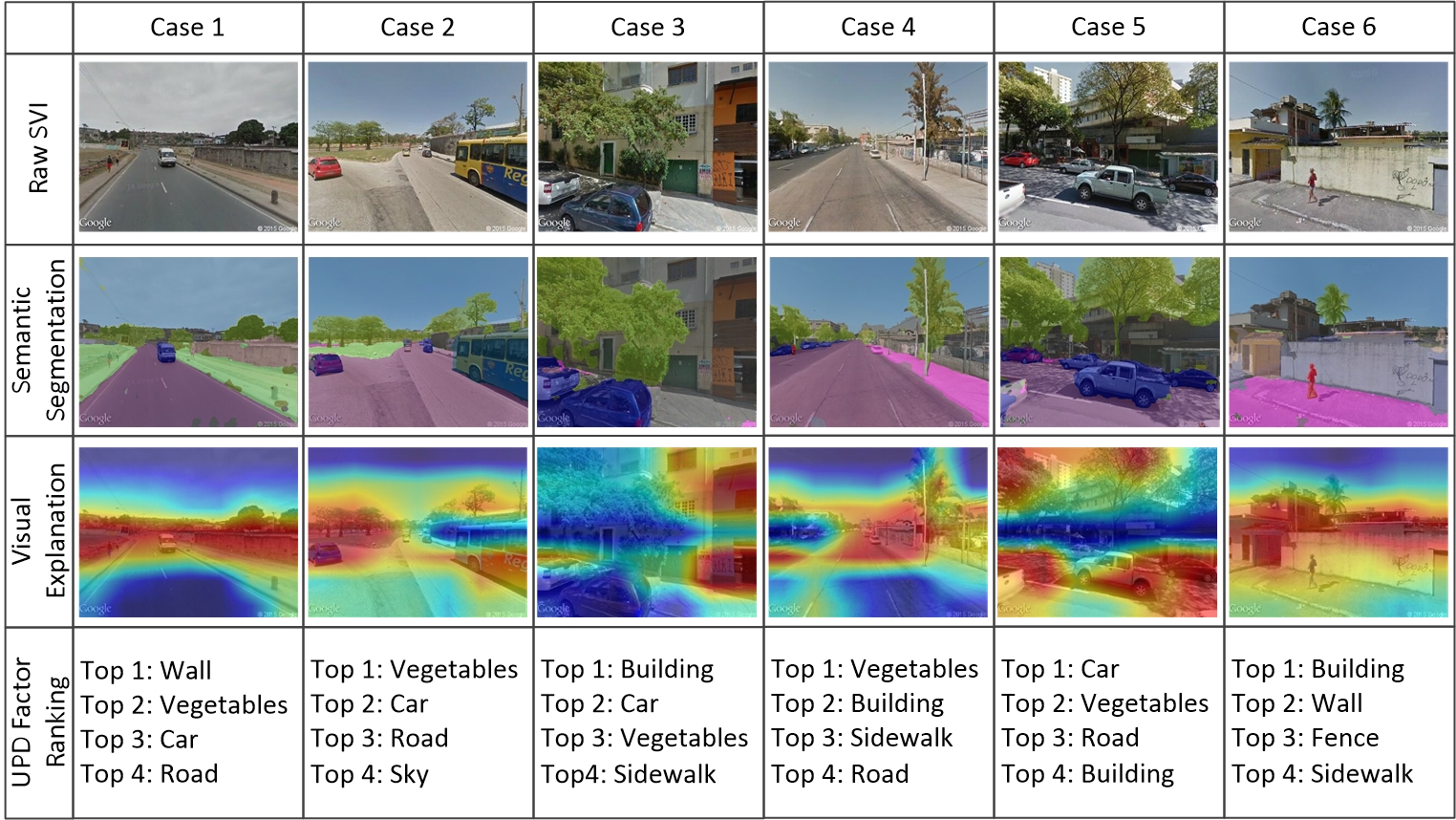}
 \caption{Examples of significant semantic objects ranked by their contribution to UPD. Note that the ranking ground truth is the same as our ranking list in the last row.}
  \label{fig:5}
\end{figure}

We also present qualitative results of the UPD ranking with several examples in Figure \ref{fig:5}. It is evident that the highest-ranking semantic objects can be precisely recognized and prioritized, and they align with our intuitive understanding of the UPD of physical disorder in urban environments. 

\subsection{Evaluation of Sensitivity to Image Segmentation and Visual Explanation Methods}

To evaluate the influence of different image segmentation and visual explanation methods on the performance of our proposed approach in UPD factor ranking, we conducted a comparative experiment using three different segmentation methods and three visual explanation methods. The segmentation methods include PSPNet \cite{zhao2017pyramid}, UNet \cite{ronneberger2015u}, and DeeLabv3 \cite{chen2017rethinking}, and visual explanation methods contain Eigen-CAM \cite{muhammad2020eigen}, Layer-CAM \cite{jiang2021layercam} and Xgrad-CAM \cite{selvaraju2017grad}.

For each combination of segmentation and visual explanation method, we used our proposed approach to rank semantic objects in street view imagery from the place pulse 2.0 dataset. We then compared the performance of our approach across the different segmentation and visual explanation methods by calculating the mAP score between our approach and ground truth annotations.

\begin{table}[t]
\begin{tabular}{l|c|c|c|c}
\hline
Methods  & Top@1 & Top@2 & Top@3 & Top@4\\
\hline
PSPNet + Score-CAM & 46.63\%  & 45.62\% &  44.90\% &  37.24\%   \\
UNet + Score-CAM & 42.65\% & 40.82\% &  38.46\% &  37.55\%  \\
DeepLabv3 + Score-CAM & 50.30\% & 48.07\% & 46.94\% & 41.84\%      \\

\hline
OCRNet + Layer-CAM & 65.19\%  & 49.57\% &  40.10\% &  28.37\%   \\
OCRNet + Eigen-CAM & 71.56\% & 67.14\% &  62.49\% &  59.49\%  \\
OCRNet + Xgrad-CAM & 58.89\% & 46.62\% & 35.00\% & 30.41\%      \\
\hline
OCRNet + Score-CAM (ours) & \textbf{87.76\%} & \textbf{82.65\%} & \textbf{78.23\%} & \textbf{75.51\%}     \\
\hline
\end{tabular}
\caption{Performance comparison (mAP) of different ablation methods in UPD factor ranking.}
\label{tab:3}
\end{table}

\begin{figure}[!h]
 \centering
 \includegraphics[width=1\linewidth]{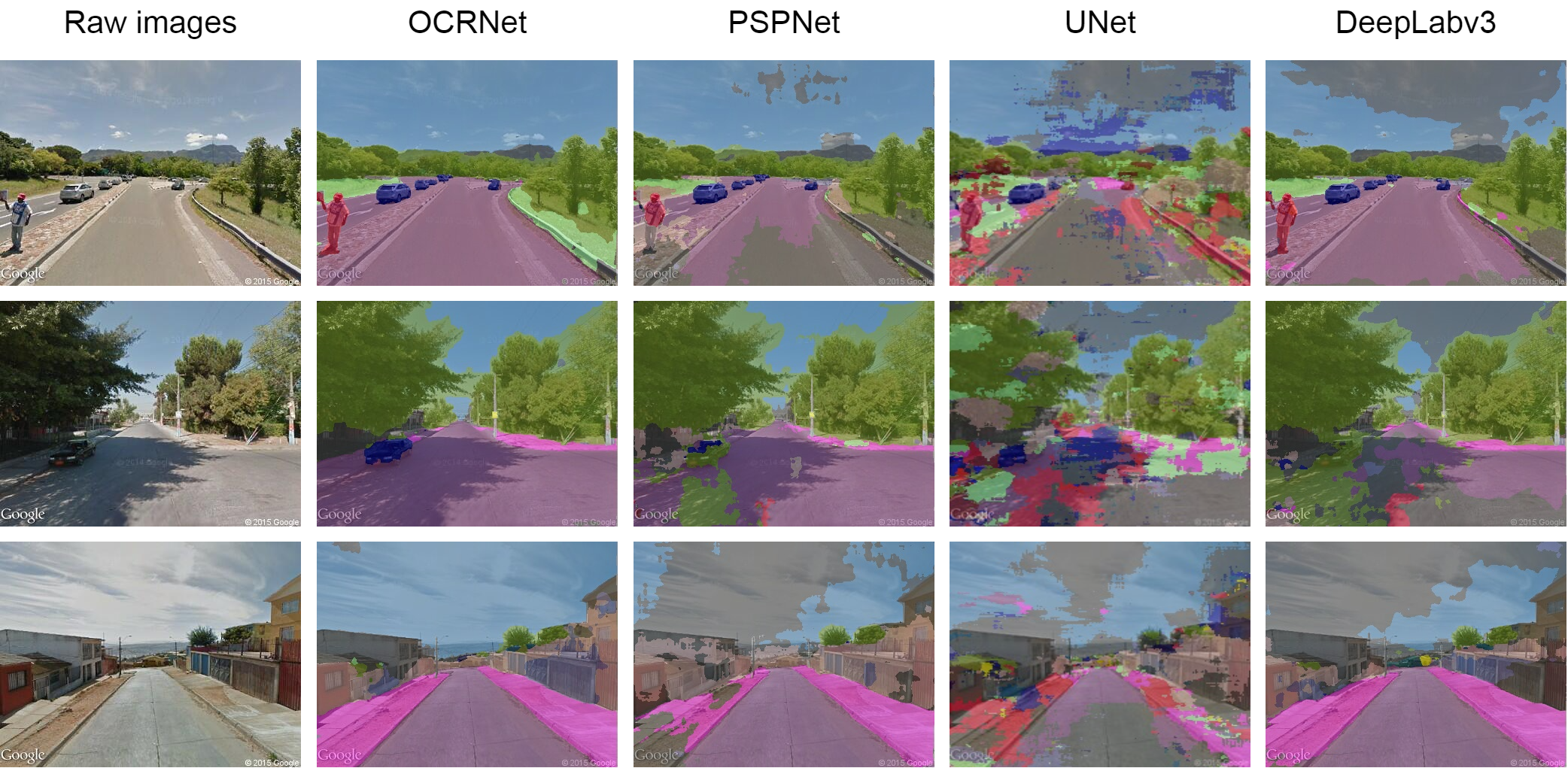}
 \caption{Qualitative performance comparison of different image segmentation methods in street view images.}
  \label{fig:6a}
\end{figure}

\begin{figure}[!h]
 \centering
 \includegraphics[width=1\linewidth]{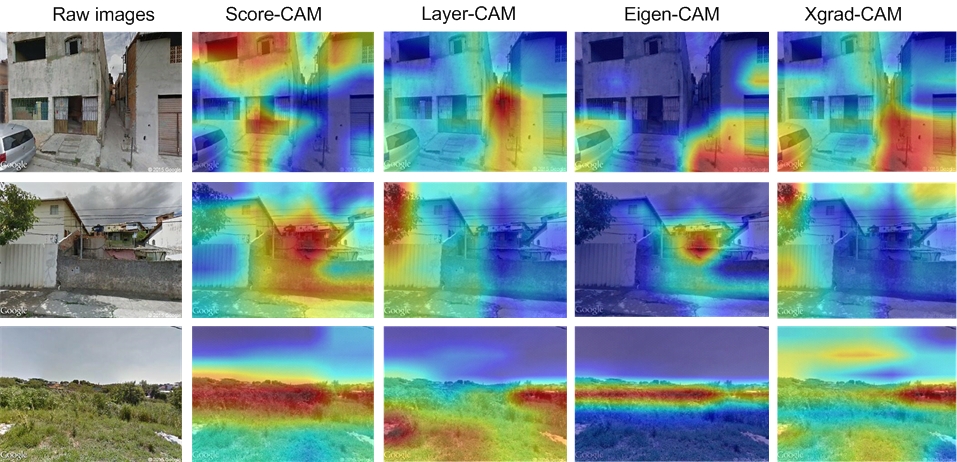}
 \caption{Qualitative performance comparison of different visual explanation methods.}
  \label{fig:6}
\end{figure}

Table \ref{tab:3} shows that our UPDExplainer with the combination of OCRNet segmentation \cite{YuanCW20} and Score-CAM~\cite{wang2020score} explanation method achieved the highest mAP scores, indicating the best performance for UPD object ranking. The sensitivity of the ranking performance is higher to semantic segmentation accuracy than to visual explanation methods, due to the significant performance variations between the four segmentation methods, as shown in Figure \ref{fig:6a}. The UNet model \cite{ronneberger2015u} with the worst segmentation results leads to the lowest UPD factor ranking scores.
The qualitative performance difference among the four visual explanation methods is qualitatively shown in Figure \ref{fig:6}. The Score-CAM method performs best in showing the density of the activation map.

These findings suggest that the choice of image segmentation and visual explanation methods may have a significant impact on the performance of UPD factor ranking. Specifically, inaccurate segmentation and visual explanation methods will lead to poor ranking performance.

\subsection{Evaluation of Sensitivity to Urban Morphology}\label{sec:4.3b}

To evaluate the generalizability of our proposed approach to detect physical disorders in different types of urban morphology \cite{hu2020classification}, we conducted an experiment to compare the performance of UPD detection across different street canyon indicators with varying $h_c/w_c$ ratios (where $h_c$ is the canyon height and $w_c$ is the canyon width). We first labeled the test data set using the $h_c/w_c$ ratio calculation method proposed in \cite{hu2020classification}. Next, we compare the performance of our approach in different urban morphologies. The results are shown in Table \ref{tab:4}.

\begin{table}[h]
\begin{tabular}{l|c|c|c|c}
\hline
Urban Morphology & Accuracy & Recall & Precision & F1         \\
\hline
$h_c/w_c=0$  & 77.46\%  & 79.63\% &  89.58\% &  84.31\%    \\
\hline
$0<h_c/w_c<1$ & 90.00\% &  91.67\% &  91.67\% &  91.67\%  \\
\hline
$1<h_c/w_c<2$ & 85.71\%   & 71.43\% & 83.33\% & 76.92\%   \\
\hline
$h_c/w_c>3$ & 83.33\% & 73.33\% &  84.62\% &  78.57 \%   \\
\hline
\end{tabular}
\caption{UPD detection performance of the proposed UPDExplainer under different street view scenarios.}
\label{tab:4}
\end{table}

It can be seen that the performance of the proposed UPDExplainer is slightly influenced by the $h_c/w_c$ ratio of the street canyon. The wide canyons of the streets, which have lower $h_c/w_c$ ratios, may provide more consistent lighting and visibility, making it easier to detect physical disorders using our proposed approach. In contrast, narrow street canyons with higher $h_c/w_c$ ratios are more difficult to detect physical disorders. 

The UPD ranking performance under different street view scenarios is shown in Table \ref{tab:5}. It demonstrates that the factors of physical disorder are more easily detected and ranked in suburban or rural areas with fewer kinds of semantic object, such as in $h_c/w_c=0$. Similarly, narrow street canyons with higher $h_c/w_c$ ratios typically have only a few main objects contributing to UPD, making it easier to rank semantic objects. Broad street canyons, on the other hand, often have a larger variety of semantic objects that contribute to UPD, making it more difficult to rank and identify significant factors.

\begin{table}[t]
\begin{tabular}{l|c|c|c|c}
\hline
Urban Morphology  & Top@1 & Top@2 & Top@3 & Top@4\\
\hline
$h_c/w_c=0$ & 91.71\%  & 89.29\% &  80.95\% &  80.80\%   \\
\hline
$0<h_c/w_c<1$ & 83.87\% & 80.00\% &  76.65\% &  70.37\%  \\
\hline
$1<h_c/w_c<2$ & 82.66\% & 76.66\% &  70.37\% &  65.69\%      \\
\hline
$h_c/w_c>3$ & 90.00\% & 86.66\% & 85.00\% & 82.50\%   \\
\hline
\end{tabular}
\caption{UPD factor ranking performance (mAP) of the proposed UPDExplainer in different street view scenarios.}
\label{tab:5}
\end{table}

These findings have important implications for the development of urban planning and policy initiatives that aim to improve physical disorders in different types of urban environments. By understanding the influence of street canyon $h_c/w_c$ ratios on the detection of physical disorders, policymakers and urban planners can design targeted interventions that are better tailored to the specific characteristics of different urban environments.

\begin{figure}[t]
 \centering
 \includegraphics[width=0.8\linewidth]{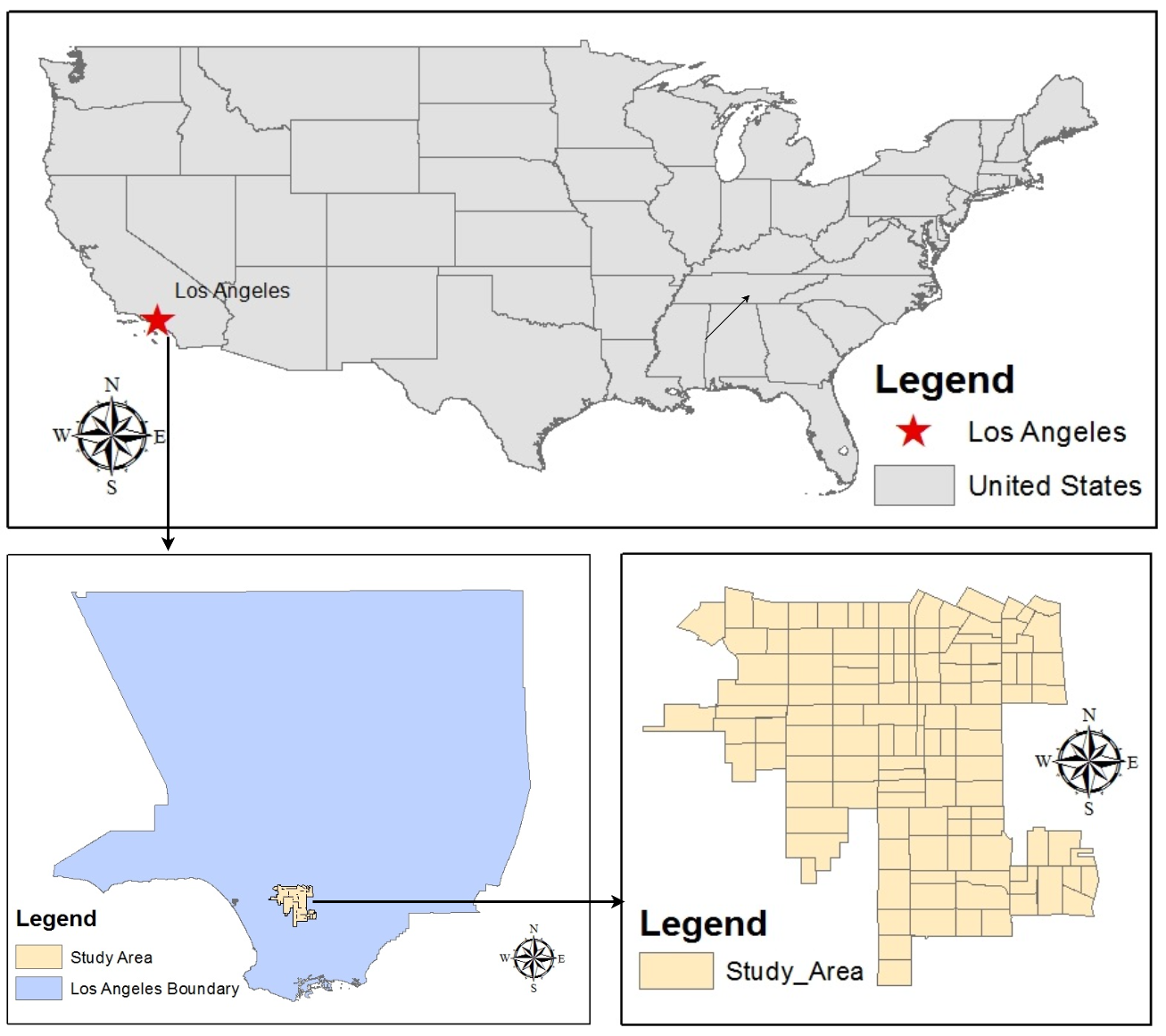}
 \caption{Geographical position of the case study area, namely the southern part of downtown Los Angeles, Los Angeles, USA, with poor street conditions.}
  \label{fig:7}
\end{figure}

\subsection{Case Study}
To demonstrate the practicability of our proposed UPDExplainer in detecting and explaining UPD, a case study was carried out in the southern region of downtown Los Angeles, California, USA (as shown in Figure \ref{fig:7}), recognized for its poor street conditions. More than 100,000 street view images were collected from various locations throughout the study area, and our approach was applied to these images for the detection of UPD and the ranking of UPD factors.

\begin{figure}[!h]
 \centering
 \includegraphics[width=1\linewidth]{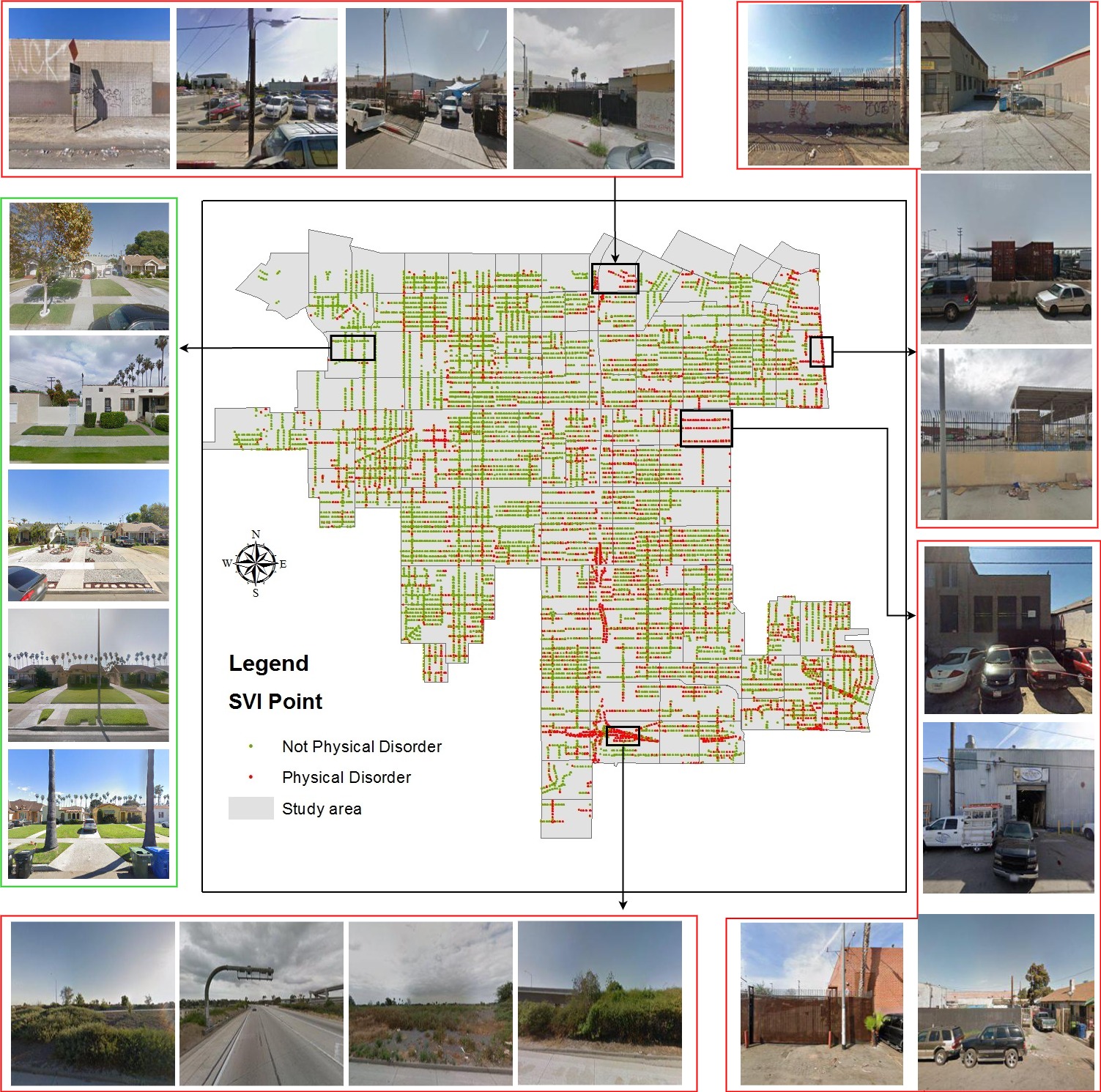}
 \caption{Mapping of UPD in the study area. We show several street view image examples with detected UPD (in red boxes) and non-UPD (in green box). Best viewed with zoom-in.}
  \label{fig:8}
\end{figure}

Figure \ref{fig:8} demonstrates the UPD distribution map detected by our proposed approach. The presence of typical disorder objects, such as litter, graffiti, and damaged infrastructure, in street view images clearly demonstrates the effectiveness of our UPDExplainer model in identifying physical disorders.

\begin{figure}[!h]
 \centering
 \includegraphics[width=1\linewidth]{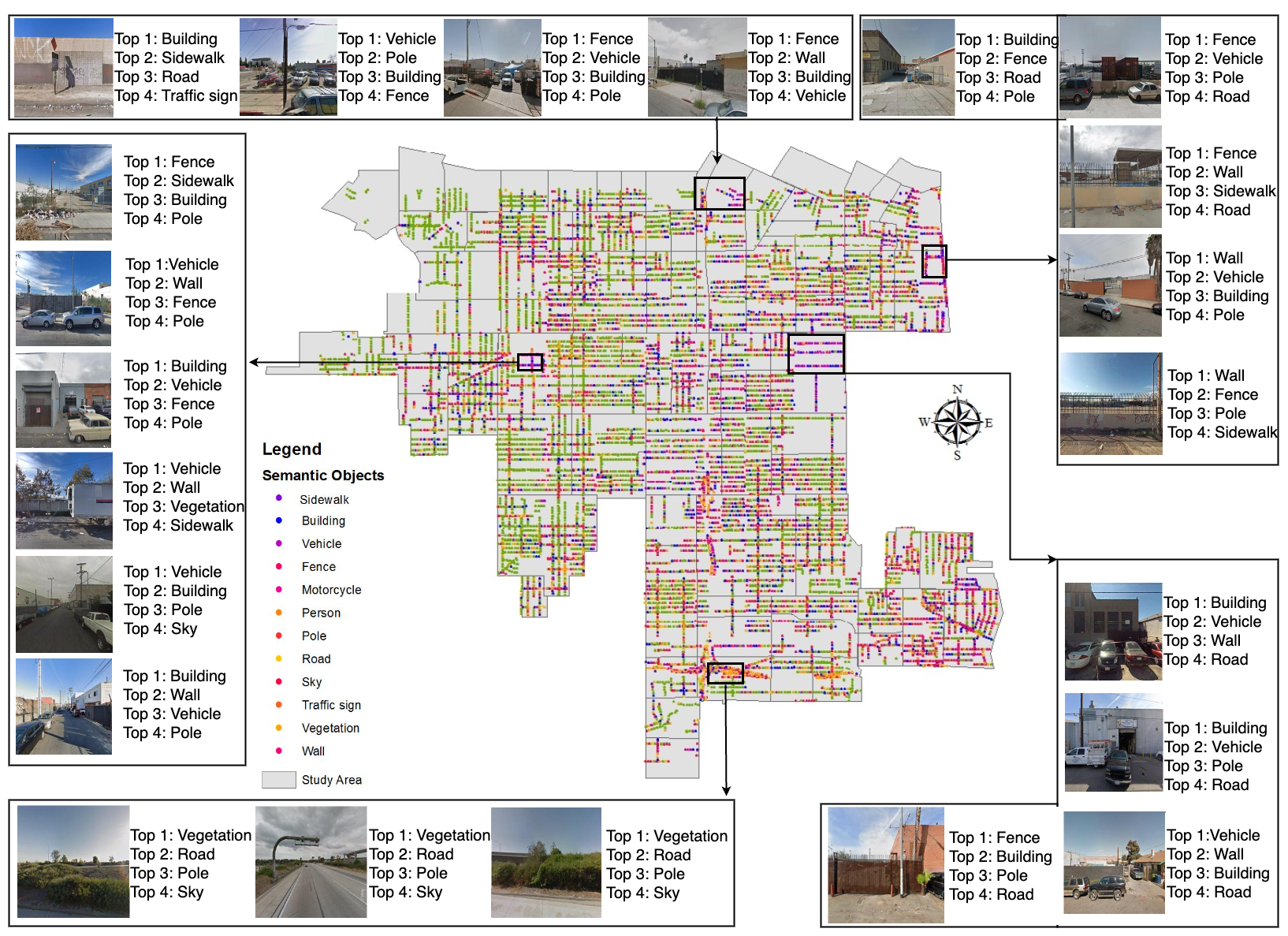}
 \caption{Mapping of identified and ranked UPD objects in the study area. We show several street view image examples with identified UPD factors. Best viewed with zoom-in.}
  \label{fig:9}
\end{figure}

Furthermore, we present the UPD factor classification results in the study case in Figure \ref{fig:9}. It is evident that our approach is capable of identifying specific semantic objects that contributed significantly to poor street conditions in the area, such as dirty sidewalks, messy vegetation, and old buildings. We believe that this will provide a clear understanding and valuable insight to improve the state of UPD in this particular area.

\section{Discussions}\label{sec:5}  
We summarize the experimental findings, followed by discussions on potential applications of the proposed UPDExplainer model. 

The evaluation results have demonstrated the superiority of the proposed UPDExplainer model over existing methods in detecting UPDs thanks to the introduction of the Swin Transformer-based model. Furthermore, the effectiveness of the proposed UPDExplainer model in identifying and ranking semantic objects in UPD has been established through extensive experimental results. Our approach, for the first time, offers a comprehensive understanding of the underlying factors that contribute to UPD.
This breakthrough has the potential to benefit the following urban planning applications.

\begin{itemize}
\item \textbf{Urban planning.} By identifying areas with high levels of physical disorder, our approach can aid urban planners and policymakers in prioritizing locations for targeted interventions to improve the physical appearance and safety of urban environments. In conjunction with ranking the semantic objects that contribute to physical disorder, our approach offers a more comprehensive understanding of the underlying causes and factors influencing physical disorder in urban settings, facilitating more informed urban planning decisions.

\item \textbf{Efficient Resource Allocation.} Our approach can also help efficiently allocate limited resources by focusing on areas with the highest levels of physical disorder.
Rapid advances in generative AI allow us to virtually allocate limited resources (e.g., via digital twins) before the deployment in the physical world.

\item \textbf{Objective Measurement}. Our approach provides an objective and quantifiable measure of physical disorder that can be used to compare and monitor changes in the physical appearance of urban environments over time. In other words, UPDExplainer can serve as a tracking tool to persistently monitor the progress in UPD reduction made by different stakeholders.

\item \textbf{Targeted Intervention Cost Balancing.} Given that the costs in terms of time and resources associated with the treatment of various factors related to UPD can differ significantly, our approach can support urban planners and policymakers in striking a balance between costs and benefits. By prioritizing areas for targeted interventions based on our model's insights, urban environments can be improved more efficiently.
\end{itemize}

\section{Conclusions}\label{sec:6}
In this article, we have presented a novel approach, named UPDExplainer, for the detection and explanation of UPD based on street view images and deep learning techniques. Our approach uses the Swin Transformer to accurately detect physical disorder in street view imagery, which also provides guidance for a novel UPD factor ranking module based on its contribution to physical disorder.

Our proposed approach makes several contributions to the field of UPD detection and explanation. First, it achieves more accurate and generalized UPD detection performance than existing methods. Second, it represents the first attempt to identify and prioritize UPD factors, offering valuable insight into the underlying causes of physical disorder. Third, it has promising practicality in real-world data applications, showing convincing results in a study area in Los Angeles, California, USA. This will potentially benefit several applications in urban planning, public policy, and community development. 

There is still potential for further improvements in our approach in future work. For example, our approach currently only uses street view imagery as input, and it may be useful to incorporate additional data sources such as satellite imagery and city maps to improve the accuracy and robustness of our approach. Furthermore, our approach focuses on physical disorder in urban environments, and it may be valuable to extend our approach to other types of environment, such as rural areas and natural landscapes.

In conclusion, our proposed approach represents an important step forward in the detection and explanation of UPD and has the potential to inform the development of effective interventions to improve the health and well-being of urban residents.

\section*{ACKNOWLEDGEMENTS}\label{ACKNOWLEDGEMENTS}

This work is partially supported by the NSF through grants CMMI-2146015 and IIS-2114644 and the WV Higher Education Policy Commission Grant (HEPC.dsr.23.7).

 \bibliographystyle{elsarticle-num} 
 \bibliography{cas-refs}





\end{document}